\pdfoutput=1

\documentclass[11pt]{article}

\usepackage[final]{acl}

\usepackage{times}
\usepackage{latexsym}
\usepackage{tabulary}
\usepackage{booktabs} 
\usepackage{amsmath}
\usepackage{mathtools}
\usepackage{amssymb}

\usepackage[T1]{fontenc}

\usepackage[utf8]{inputenc}

\usepackage{microtype}

\usepackage{inconsolata}

\usepackage{graphicx}

\title{REXEL: An End-to-end Model for Document-Level Relation Extraction and Entity Linking}

\author{Nacime Bouziani\thanks{Work completed whilst at Amazon Alexa AI}$^{1}$, Shubhi Tyagi$^{2}$, Joseph Fisher$^{2}$, Jens Lehmann$^{2}$, Andrea Pierleoni$^{2}$ \AND
  \normalfont{$^{1}$I-X Centre for AI In Science} \\ 
  Imperial College London, London, UK \And
  \normalfont{$^{2}$Amazon Alexa AI} \\
  Cambridge, UK\AND\vspace{-0.6cm}
  \\ 
  \texttt{n.bouziani18@imperial.ac.uk} \\
  \texttt{\{tshubhi, fshjos, jlehmnn, apierleo\}@amazon.com}
  }

\begin{document}
\maketitle
\begin{abstract}

Extracting structured information from unstructured text is critical for many downstream NLP applications and is traditionally achieved by \textit{closed information extraction} (cIE). However, existing approaches for cIE suffer from two limitations: \textit{(i)} they are often pipelines which makes them prone to error propagation, and/or \textit{(ii)} they are restricted to sentence level which prevents them from capturing long-range dependencies and results in expensive inference time. We address these limitations by proposing REXEL, a highly efficient and accurate model for the joint task of document level cIE (DocIE). REXEL performs mention detection, entity typing, entity disambiguation, coreference resolution and document-level relation classification in a single forward pass to yield facts fully linked to a reference knowledge graph. It is on average 11 times faster than competitive existing approaches in a similar setting and performs competitively both when optimised for any of the individual sub-tasks and a variety of combinations of different joint tasks, surpassing the baselines by an average of more than 6 F1 points. The combination of speed and accuracy makes REXEL an accurate cost-efficient system for extracting structured information at web-scale. We also release an extension of the DocRED dataset to enable benchmarking of future work on DocIE, which is available at \url{https://github.com/amazon-science/e2e-docie}.
\end{abstract}

\vspace{-4mm}
\begin{figure*}[htp]
\centering
\includegraphics[width=\textwidth]{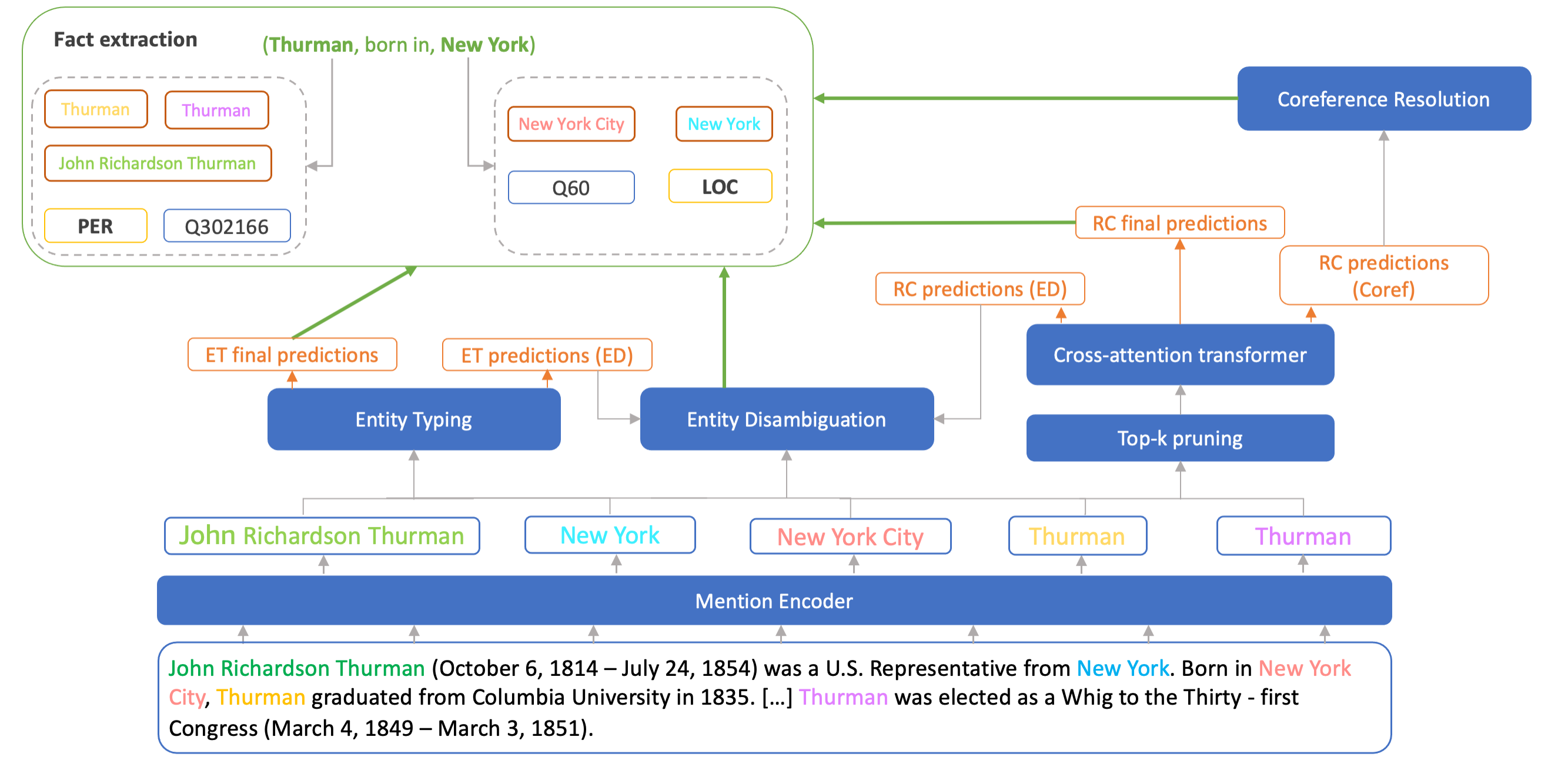}
\caption{REXEL model architecture illustrating the interaction between different components. The model takes the raw text as input and yields fully linked facts expressed across a document.}
\label{fig:E2E_model_architecture}
\vspace{-4mm}
\end{figure*}

\section{Introduction}
\label{sec:Intro}
\vspace{-2mm}
Extracting structured information from unstructured text is a critical step for many downstream NLP tasks like knowledge graph construction \cite{muhammad_open_2020}, question answering \cite{yao_information_2014}, knowledge discovery \cite{trisedya_neural_2019}, and text summarization \cite{genest2012fully}. In cIE, this is defined as extracting an exhaustive set of (subject, relation, object) triples, or \textit{facts}, from unstructured text that are \textit{fully linked}, i.e., consistent with a predefined set of entities and relations from a knowledge graph (KG) schema. cIE can be further decomposed into the subtasks: \textit{mention detection} (MD), \textit{entity typing} (ET), \textit{entity disambiguation} (ED), and \textit{relation classification} (RC).

Traditionally, cIE is done by combining these subtasks sequentially \cite{nasar_named_2021}, which involves the use of separate and often different models for each task to yield facts that can be ingested into a KG. However, such pipeline architectures are prone to error accumulation from each component leading to significant deterioration of the overall performance~\cite{miwa_modeling_2014, trisedya_neural_2019, mesquita_knowledgenet_2019}. Additionally, pipeline architectures assume a one-way dependency between the subtasks, disregarding the dependencies among components that could effectively boost performance. For instance, while ED typically informs RC, recent works have demonstrated that RC information can also be effectively utilised for the ED task~\cite{KBED}, and help preventing issues such as popular entities overshadowing less common entities~\cite{shadowlink}. Consequently, various joint/end-to-end (E2E) systems have been proposed to address this issue by jointly performing NER and RC \cite{miwa_modeling_2014, pawar_end--end_2017}. This joint task is often referred to as \textit{relation extraction} (RE). However, these approaches do not address ED and thus do not yield facts fully linked to a KG.

Another drawback of existing approaches for cIE is that they mostly operate at sentence level, i.e., perform RC between two entities from a single sentence at a time~\cite{cai_bidirectional_2016, han_hierarchical_2018, feng_reinforcement_2018}. Thus, they capture limited sentence-level context and miss the facts that are expressed between entities across sentences. This severely limits the amount of information that can be extracted from the web. According to ~\cite{yao_docred_2019}, 40.7\% of the facts in a document can only be determined at the document level. Also, sentence-level approaches require a forward pass for each sentence, often leading to higher inference times, which makes them inefficient for web-scale applications. In contrast, document-level RC is computationally more efficient as it extracts triples over an entire document in a single forward pass. To address these issues, several models have been proposed for document-level RC ~\cite{zeng_double_2020, wang_global--local_2020, xu_entity_2021, zhang_document-level_2021} but they do not perform the remaining sub-tasks needed for DocIE.

To address the above problems we introduce \textbf{REXEL}, a computationally efficient E2E model for DocIE. REXEL takes unstructured text and extracts facts which are fully linked to a reference KG in a single forward pass per document. It has a modular architecture in which the various subtasks for DocIE inform each other by leveraging intermediate embedding representations. Thus, the proposed framework facilitates deployment not only for DocIE but also for various combinations of its 5 subtasks (e.g., use MD and ET only for NER). The combination of modularity, fast inference speed and high accuracy makes REXEL suitable for performing DocIE or its sub-tasks at industry scale.\\

To summarize, our contributions are as follows:
\vspace{-2mm}
\begin{enumerate}
    \item We introduce REXEL, a unified E2E model for DocIE, i.e., extracting facts at document level fully linked to a reference KG in a single forward pass per document.
    \vspace{-2mm}
    \item We demonstrate that though REXEL is optimised for the E2E task of DocIE, it maintains a competitive edge with related work in E2E RE setting and all its individual subtasks. Specifically, REXEL improves upon the baselines for the E2E RE task by an average of >6 F1 points across datasets. When comparing the performance of individual subtasks, we observe that REXEL outperforms the baselines by an average of 6 F1 points.
    \vspace{-2mm}
    \item We also demonstrate that when compared to other E2E RE models, in the same setting REXEL is on average 11 times faster.
    \vspace{-2mm}
    \item Finally, we release an extension of the DocRED \cite{yao_docred_2019} dataset released by~\cite{eberts_end--end_2021} augmented with silver standard labels for entity linking to facilitate benchmarking of future work on DocIE. We name this extension \textit{DocRED-IE}.
\end{enumerate}

\section{Related work}

\subsection{Closed Information Extraction (cIE)}
\vspace{-2mm}
Several E2E systems have been proposed for cIE \cite{liu_seq2rdf_2018, trisedya_neural_2019, sui_set_2021, josifoski-etal-2022-genie}. However, all these methods are sentence-level architectures and therefore they inherently lose triples expressed across sentences. They are also prohibitively expensive for deployment at web-scale since the inference compute increases linearly with the number of sentences requiring a forward pass for each sentence. 

In comparison, DocIE is a significantly more challenging task as it involves capturing long-range dependencies effectively to extract relations between entities which are further apart from each other in the text. Scaling cIE to document level from sentence level also requires an additional subtask of \textit{coreference resolution} (Coref), i.e., group all the different mentions in the document referring to the same entity.

\subsection{Document-level Relation Extraction}
\vspace{-2mm}
Various E2E models have been proposed that combine the task of NER and document-level RC in a joint setting \cite{eberts_end--end_2021, zaporojets_dwie_2021}. Other works such as REBEL \cite{huguet-cabot-navigli-2021-rebel-relation} and KBIE \cite{verlinden_injecting_2021} have proposed using additional data like the Wikipedia text, hyperlinks and Wikidata KG to further improve RE performance.
However, these approaches do not perform ED and hence do not yield facts fully linked to a reference KG. Thus, ingesting the output of such models in a KG necessitates a separate ED model to link the extracted entities. This again results in a pipeline architecture between RE and ED models.

To the best of our knowledge, REXEL is the first E2E model to extract facts which are fully linked to a reference KG, at document level and address the task of DocIE. Also, while relation classification (RC) is also usually referred to as \textit{relation extraction} (RE), the E2E literature has adopted different conventions. For sake of consistency with prior works \cite{eberts_end--end_2021, miwa_end--end_2016}, we use RC to refer to the extraction of relations between entity pairs and RE to refer to the E2E task including MD, ET, RC, and Coref. 

\section{REXEL}
\vspace{-2mm}
We introduce REXEL (\textbf{R}elation \textbf{Ex}traction and \textbf{E}ntity \textbf{L}inking), a novel end-to-end model for DocIE. REXEL extracts triples fully linked to a KG by jointly performing MD, ET, document level RC, Coref and ED in a single forward pass. It combines the 5 subtasks in a unified architecture via intermediate embedding representations. This facilitates each task to inherently benefit from each other, significantly boosting task accuracy, extracting facts expressed across sentences, and maintaining computational efficiency for web-scale deployment. Figure \ref{fig:E2E_model_architecture} illustrates the architecture and each module is detailed in following sections.

\subsection{Task Formulation}
\vspace{-2mm}
Given a KG with a set of entities $E=\{e_1, e_2, \dots, e_{|E|}\}$, entity types $T=\{t_1, t_2, \dots, t_{|T|}\}$, and relations $R=\{r_1, r_2, \dots, r_{|R|}\}$, let $X=\{x_1, x_2, \dots, x_{|X|}\}$ be the sequence of tokens in a document(d). The goal of DocIE is to extract linked facts, i.e., $\mathcal{G}: X \rightarrow G$ with $G \subseteq E \times R \times E$ being a set of triples. 
This is done by \emph{(i)} \textbf{MD}: extracting mention spans resulting in a list of subsets of $X$, 
\emph{(ii)} \textbf{Coref}: clustering mentions into entities, 
\emph{(iii)} \textbf{ET}: extracting the entity types for each cluster, 
\emph{(iv)} \textbf{RC}: extracting relations by mapping entity pairs $\{e_1, e_2\}$ to relations $r \in R$ and
\emph{(v)} \textbf{ED}: assigning each cluster of mentions to a corresponding KG entity $e \in E$.

\subsection{Mention Detection (MD)}
\label{sec:MD}
\vspace{-2mm}
We encode the tokens $x_i$ in the input text document using RoBERTa~\cite{liu_roberta_2019} and use the contextualised token embeddings $\mathbf{h_i}$ from the final layer of the encoder for the token $x_i$. The tokens are encoded using the BIO tagging format \cite{ramshaw-marcus-1995-text}. We then train a linear layer to perform token classification from the token embeddings $\mathbf{h_i}$ using cross-entropy loss $\mathcal{L}_m$ with respect to the gold token labels. We obtain mention embeddings $\mathbf{m_i}$ for each mention $m_i$ by average pooling the contextualised token embeddings ($\mathbf{h_i}$) for all tokens in a mention from the final transformer layer. The output of this module is a list of mention spans present in the input text along with their contextualised embeddings.

\subsection{Entity Typing (ET)}
\label{sec:ET}
\vspace{-2mm}
Given a fixed set of types $t \in T$, the ET module is trained by applying a linear layer $f_1$ followed by a sigmoid activation to the mention embedding $\mathbf{m_i}$ to predict an independent unnormalised score for each type $t$ for each mention $m_i$.  
REXEL produces two independent predictions for ET. 
The $ET_{ed}$ layer predicts fine-grained Wikidata types (1.3k) that are later used to inform ED. We do not train on this explicitly, but via ED (see Section \ref{sec:ED}). The $ET_{final}$ layer predicts the type(s) for each mention according to the ones permissible within the target dataset for the target task. We train this module from the gold entity types using binary cross-entropy loss $\mathcal{L}_{t}$ corresponding to $ET_{final}$ predictions. 
There are two separate predictions for ET as the target dataset may not have as many or the same fine grained types. Fine grained entity types provide critical additional information that can inform ED and thus boost overall performance. We aggregate predictions at entity cluster level by selecting the most frequent type among the cluster mentions as the entity type. REXEL supports both single and multiple type classification.

\subsection{Relation Classification (RC)}
\label{sec:RE}
\vspace{-2mm}
REXEL extracts relations at mention-level using a cross-attention transformer and uses the coreference resolution predictions to map the extracted relations to the entity-level. We employ top-k pruning from~\cite{lee_higher-order_2018} to extract relations only for the $k$ mention pairs with highest probabilities of being connected by a relation. This probability is computed for each mention pair using a bilinear layer. This first stage results in less accurate but more efficient predictions and is referred to as the \textit{coarse stage}. However, in REXEL the coarse stage is adopted for both: relation classification and coreference resolution. The coarse stage is then followed by the \textit{fine stage}, which extracts relations between surviving mention pairs. The resulting coarse-to-fine RC module yields competitive accuracy with high efficiency. Similar to the ET module, we have multiple prediction layers for RC: $RC_{ed}$, which predicts the Wikidata relations and is used as an input to the ED module, $RC_{coref}$, which predicts the pairwise coreference scores for the Coref module, and $RC_{final}$, which is the final prediction layer on the target relations of the given dataset. This module is trained from the gold mention spans, gold entity types, gold entity IDs and gold clusters using binary cross-entropy loss $\mathcal{L}_{r}$ with respect to the gold triples on the $RC_{final}$ prediction layer only.

\subsection{Coreference Resolution (Coref)}
\label{sec:Coref}
\vspace{-2mm}
This module has two stages: the first predicts pairwise coreference scores for each mention pair that remains after top-k pruning, and the second uses pair predictions to form entity clusters by using average linkage clustering based on a given distance threshold. Other approaches like greedy clustering, complete linkage and clustering via Wikidata identifiers resulted in similar performance. More details can be found in Appendix \ref{sec:coref_app}. The first stage can be expressed as a relation classification task with one relation that determines whether two mentions are coreferent to each other. Hence, we delegate this stage to the RC cross-attention transformer. The training is done with respect to the predicted coreference scores only. We train this module from the pairwise scores of the gold mention spans using binary cross-entropy loss $\mathcal{L}_{c}$ with respect to the gold clusters. The output of this module is a group of entity clusters in a document and their corresponding mentions.

\subsection{Entity Disambiguation (ED)}
\label{sec:ED}
\vspace{-2mm}
REXEL links each entity mention in the text to a unique Wikidata ID using a training procedure similar to~\cite{KBED}. The ED module takes as input the mention embeddings $\mathbf{m_i}$, entity type predictions for ED $\mathbf{t_{ed}}$ and RC predictions for ED $\mathbf{r_{ed}}$. We also add a global entity prior $\hat{P}(e|m)$ (PEM score), which is the probability of an entity given the mention text and is obtained from hyperlink count statistics as done in \cite{Raiman2018DeepTypeME}. We train this module from gold mention spans and gold entity types $\mathbf{t_{ed}}$ by using binary cross-entropy loss $\mathcal{L}_{d}$ with respect to the gold entity IDs. Note that we do not train on $ET_{ed}$ and $RC_{ed}$ explicitly, instead, the training for those predictions is done using the signal from $\mathcal{L}_{d}$ only. REXEL performs ED for each mention and we get the entity IDs at the cluster level (i.e., when multiple mentions are clustered together by coref) by taking the majority vote of the entity IDs for all the mentions in the cluster.

\subsection{Optimization and Inference}
\label{sec:Optimization}
\vspace{-2mm}
REXEL is optimised using a weighted sum of the module-specific losses with fixed weights, which are tunable hyperparameters as follows: 

\begin{equation}
    \label{eq:total_loss}
	\mathcal{L} = \lambda_1 \mathcal{L}_m + \lambda_2 \mathcal{L}_t + \lambda_3 \mathcal{L}_d +  \lambda_4 \mathcal{L}_c +  \lambda_5 \mathcal{L}_r
\end{equation}

When training on a single subtask, the weights for all the other task losses are set to zero. When training for the RE task, $\lambda_3$ is set to zero. For individual subtask inference, we use gold labels for the other tasks. For the RE inference, we use the predicted mention spans, predicted entity types, predicted coref clusters and predicted entities as input. Training environment details are in Appendix \ref{sec:training-details}.

\section{Experiments}
\label{sec:Experiments}
\vspace{-2mm}
\subsection{Datasets}
\vspace{-1mm}
We report performance on DWIE~\cite{zaporojets_dwie_2021}, the only dataset available supporting DocIE. We also augment the end-to-end DocRED split (DocRED-E2E) ~\cite{eberts_end--end_2021}, which does not support annotations for ED, with silver annotations for entity links, and release the resulting dataset for future works. For this, we use the SoTA EL model ReFinED~\cite{refined} to link the mention spans against Wikidata and report DocIE performance on the DocRed-E2E split augmented with these entity links.
We also report performance on DocRED-E2E for the E2E RE task, which allows comparison with existing approaches. More details on the datasets can be found in Appendix \ref{sec:datasets_appendix}.

\begin{table*}[ht]
    \centering
    \small
    \begin{tabular}{lllrrrrrrr}
    \toprule
    \textbf{Training Setup} & \textbf{Dataset} & \textbf{Model} & \multicolumn{6}{c}{\textbf{Subtasks}} & \textbf{E2E} \\
    \cmidrule{4-9}
    & & & \textbf{MD} & \textbf{ET} & \textbf{NER} & \textbf{ED} & \textbf{Coref} & \textbf{RC} &\\
    \midrule
     \textbf{Subtask}     & DWIE               & DWIE & N/A & N/A & 87.1 &  N/A  & \underline{91.1} & \underline{71.3} & N/A \\
                        & & REXEL & 96.37 & 93.53 & N/A & 93.22 & \textbf{96.05} & \textbf{74.89} & N/A\\
    \cmidrule{2-10}
                        
    & DocRED  & JEREX & \textbf{92.66} & \underline{95.29} & N/A& N/A &  \underline{90.46} & \underline{59.76} & N/A  \\
                             & & REXEL & \underline{90.56} & \textbf{96.01}  & N/A & 86.74 & \textbf{90.93} & \textbf{60.10} & N/A \\

    \midrule
    \textbf{RE} & DWIE  & DWIE & N/A & N/A & \underline{88.8} &  N/A  & \underline{91.6} & N/A & 50.4  \\
                       & & KBIE & N/A & N/A & 75 &  N/A  & 91.5 & N/A & \underline{52.1} \\
                        & & REXEL & 95.88 & 93.00 & \textbf{90.59} & N/A & \textbf{95.12} & 68.3 & \textbf{65.8} \\
     \cmidrule{2-10}
                        
    & DocRED  & JEREX & \textbf{92.99} & \underline{80.10} & N/A & N/A & 82.79 & N/A & \textbf{40.38} \\
                            & & KBIE & N/A & N/A & \underline{71.8} & N/A & \underline{83.6} & N/A & 25.7 \\
                            & & REXEL & \underline{90.68} & \textbf{95.78} & \textbf{87.49} & N/A & \textbf{89.02} & 57.38 & \underline{39.06} \\
    
    \midrule
    \textbf{DocIE}  & DWIE      & REXEL & \textbf{95.35} & \textbf{92.76} & \textbf{89.39} & \textbf{91.19} & \textbf{93.01} & \textbf{62.04} & \textbf{53.77} \\
    
    & DocRED* & REXEL & \textbf{90.1} & \textbf{95.63} & \textbf{86.19} & \textbf{86.23} & \textbf{86.59} & \textbf{53.63} & \textbf{27.96}\\
    \bottomrule
    \end{tabular}
    \caption{Model evaluations under various training setups evaluated individually on each subtask and the end-to-end (E2E) task. N/A denotes that the model does not support evaluation for that task. The best performing models are marked in \textbf{bold} and the second best are \underline{underlined}. For DocIE training, we report the first numbers for the two datasets. * DocRED end to end split augmented with ReFinED \cite{refined} entity links.}
    \label{tab:Individual_tasks_DocRED}
\vspace{-4mm}
\end{table*}

\subsection{Evaluation settings}
\vspace{-1mm}
\subsubsection{Subtask}
\vspace{-1mm}

In the \textit{Subtask} training setting, we train and evaluate each of the 5 DocIE subtasks independently as mentioned in \ref{sec:Optimization}. This setting measures the ceiling performance of each component. We report these metrics to understand the impact of the performance of each component as we move from independent subtask training to E2E RE and E2E DocIE training settings.

\subsubsection{Relation Extraction (RE)}
\vspace{-1mm}
 Despite the recent works on the joint entity and relation extraction task for document-level RE, there has been a lack of a cohesive task definition and consistent baselines, leading to discrepancies in dataset usage and evaluation procedures, as discussed in~\cite{taille-etal-2021-separating}.
We follow the \textit{hard-metric} setting to evaluate the E2E RE task in line with previous works~\cite{eberts_end--end_2021, zaporojets_dwie_2021}. More precisely, a triple is considered as correct if the relation type and the entity clusters associated to the head and tail entities are correct. An entity cluster is correct if the clustered mentions and the entity type match a ground truth entity cluster. Finally, a mention is correct if it matches exactly a ground truth mention span. This evaluation setting penalizes clustering mistakes, i.e., if a given predicted entity cluster is incorrect, all the gold triples associated with all the gold entity clusters which have at least one mention span belonging to that predicted entity cluster will not be resolved correctly. Other metrics have been proposed to alleviate the constraint on predicted clusters, such as the \textit{soft metric} in~\cite{zaporojets_dwie_2021}.

While DocRED is restricted to one type per entity, DWIE allows multiple types per entity. Hence, for DWIE we aggregate mention-level predictions to form the entity-level types predictions by taking the union of the predicted types of the mentions in the cluster in agreement with previous work~\cite{zaporojets_dwie_2021, verlinden_injecting_2021}.

\subsubsection{Document level closed Information Extraction (DocIE)}
\vspace{-1mm}
As document-level RE does not link entities, we extend the evaluation setting to address the joint DocIE task. We introduce the \textit{DocIE hard metric} for the E2E task: A triple is correct if the relation type and the entity clusters associated with the head and tail entities are correct. An entity cluster is correct if the clustered mentions, the entity type and the entity identifier match a ground truth entity cluster. Finally, a mention is correct if it matches exactly a ground truth mention span.

\subsubsection{Inference Speed}
\vspace{-2mm}
Since we are pioneering the task of DocIE, we do not have a related work to compare REXEL's performance in this setting. Thus, we compare REXEL's inference speed with JEREX~\cite{eberts_end--end_2021} and DWIE \cite{zaporojets_dwie_2021} in the RE setting. We use the code released by the authors to report the inference time. Both of these works support inference only for their respective datasets, i.e., DocRED-E2E and DWIE respectively.

\section{Results}
\vspace{-2mm}
We summarize all results from single runs in Table \ref{tab:Individual_tasks_DocRED}. Note that DWIE and KBIE \cite{verlinden_injecting_2021} report performance on NER instead of MD and ET separately. Therefore, they are only comparable for Coref and RC in the subtask setting. In E2E RE and E2E DocIE settings, we also report REXEL's performance on NER, which requires both the mention span and the entity type to be correct. We follow~\cite{zaporojets_dwie_2021} for the scoring mechanism for evaluating NER performance. We demonstrate that the performance of REXEL on joint tasks (RE and DocIE) is on par with task-specific learning, while being more efficient due to shared parameters and training steps. 
\subsection{Subtask}
\vspace{-1mm}
In order to assess the performance of each component of REXEL, we train and evaluate each subtask individually on DWIE and DocRED-E2E split. 
When trained on individual subtasks only, REXEL improves upon the SoTA model on DWIE by an average of 4 F1 points while surpassing the SoTA on DocRED-E2E on all subtasks except MD. Note that JEREX and DWIE are not only the SoTA in the RE setting but also in the subtask setting.
\vspace{-1mm}
\subsection{Relation Extraction (RE)}
\vspace{-2mm}
In the E2E RE setting, we compare with three other related works: JEREX~\cite{eberts_end--end_2021}, DWIE~\cite{zaporojets_dwie_2021} and KBIE~\cite{verlinden_injecting_2021}. JEREX and DWIE report performance on DocRED-E2E and the DWIE dataset for RE, as well as performance on each subtask, thus being directly comparable with our setting. On the other hand, KBIE only reports performance when trained for the E2E task. We do not compare with REBEL~\cite{huguet-cabot-navigli-2021-rebel-relation} since their E2E evaluation is less strict and thus is not a fair comparison to JEREX and REXEL \footnote{https://github.com/lavis-nlp/jerex/issues/15}. 

We find that REXEL outperforms the baselines on DWIE for all the individual subtasks and improves upon the SoTA on the E2E RE task by almost 14 F1 points. However, on DocRED-E2E even though REXEL improves upon JEREX for the subtasks by an average of >6 F1 points the improvement does not translate into a corresponding boost in E2E RE task. This can be attributed to the false negatives prevalent in the dataset (64.6\%), which penalize the model due to missing annotations~\cite{tan2022revisiting}, significantly hampering the E2E hard metric. Also, while subtask training setting involves a single task-specific loss, the E2E RE setting involves multiple losses (cf. equation \eqref{eq:total_loss}), which dilutes the training effort over all the subtasks. This explains the slight drop in the subtasks' performance when comparing models trained in the E2E RE setting against models trained in the subtask setting. However, the E2E approach yields better E2E performance than the pipeline approach as it does not suffer from the propagation of errors.
\vspace{-1mm}
\subsection{Document level closed Information Extraction (DocIE)}
\vspace{-2mm}
For both datasets, we observe that REXEL is able to scale from the E2E RE to E2E DocIE by incorporating ED. For all the subtasks we observe comparable performance between models trained for RE and DocIE, indicating that adding ED to the joint task setting does not deteriorate REXEL's performance on individual subtasks. 

On the other hand, we observe a significant drop in the E2E task because of the additional criterion in the proposed hard metric for DocIE. In this setting, a cluster is considered incorrect if its corresponding entity identifier is incorrect, thus all the triples extracted for such a cluster are considered incorrect.

\subsection{Inference Speed}
\vspace{-2mm}

We report the comparison of inference speed across datasets in Table~\ref{tab:Speed}. REXEL is on average almost 11 times faster than the baselines (19x on DocRED and 3x on DWIE) in the E2E RE setting, i.e., without performing ED. This can be explained by our coarse-to-fine approach, which reduces training/inference time while still preserving competitive accuracy. Even in the E2E DocIE setting, REXEL remains faster than the baselines while performing the additional task of ED. 

\begin{table}[h]
\vspace{-3mm}
    \centering
    \small
    \begin{tabular}{lcc}
    \toprule
     &  \textbf{DocRED} & \textbf{DWIE} \\
    \midrule
    \textbf{JEREX} & 344 & N/A\\
    \textbf{DWIE} & N/A & 82 \\
    \textbf{REXEL (RE)} & \textbf{18} & \textbf{27} \\
    \textbf{REXEL (DocIE)} & \underline{90} & \underline{74}\\
    \bottomrule
    \end{tabular}
    \caption{Inference speed comparison in seconds. The best values are in \textbf{bold} and the second best are \underline{underlined}. N/A denotes that the code release does not support inference on the target dataset.}
    \label{tab:Speed}
    \vspace{-6mm}
\end{table}

\section{Conclusion}
\vspace{-2mm}

In this work we introduce REXEL, a highly efficient and accurate end-to-end model for document-level closed information extraction. REXEL extracts facts from unstructured text which are fully linked to a reference KG for an entire document in a single forward pass. We further demonstrate that REXEL is 11 times more computationally efficient than baselines in the same setting, while improving upon the existing baselines on E2E RE by an average of 6 F1 points across datasets and across different task settings. Specifically, we improve upon the state-of-the-art on DWIE for E2E RE by almost 14 F1 points.  We report the first numbers for DocIE on DWIE and DocRED-E2E augmented with entity links. We also release the latter dataset to facilitate benchmarking of future works on DocIE. Thus, the combination of accuracy, speed and scale makes REXEL suitable for
being deployed to extract fully linked facts from web-scale unstructured data with state-of-the-art accuracy and an order of magnitude lower cost than existing approaches.

\section*{Limitations}
One limitation of our work is that REXEL currently supports fact extraction for entities only and will miss the facts for relations where either the subject or object is a string literal. We leave the extension of REXEL to extract string literal-based facts for future work. Another limitation is that, for a given document, the context length of REXEL is limited to the maximum number of tokens that can be encoded by the base transformer, which is RoBERTa \cite{liu_roberta_2019} in our case (see Section \ref{sec:MD}). This implies that the model cannot capture triples that involve very long-range dependencies that go beyond the maximal context length. In practice, we find that this problem is negligible in our case as only a few triples fall into that category for both DocRED and DWIE. However, this might have a stronger impact for other applications. In addition, this limitation is not specific to the REXEL architecture per se but is inherent to the transformer used. Finally, while the proposed DocIE hard metric provides a common ground for future benchmarks on DocIE, it may not fully align with some industrial applications where missing a few mentions within entity clusters is not critical. In such contexts, the hard metric would provide a lower bound on the performance, and other metrics can be considered for better alignment with specific application requirements.

\bibliography{anthology,custom}

\begin{thebibliography}{37}
\expandafter\ifx\csname natexlab\endcsname\relax\def\natexlab#1{#1}\fi

\bibitem[{Ayoola et~al.(2022{\natexlab{a}})Ayoola, Fisher, and Pierleoni}]{KBED}
Tom Ayoola, Joseph Fisher, and Andrea Pierleoni. 2022{\natexlab{a}}.
\newblock \href {https://doi.org/10.18653/v1/2022.naacl-main.210} {Improving entity disambiguation by reasoning over a knowledge base}.
\newblock pages 2899--2912.

\bibitem[{Ayoola et~al.(2022{\natexlab{b}})Ayoola, Tyagi, Fisher, Christodoulopoulos, and Pierleoni}]{refined}
Tom Ayoola, Shubhi Tyagi, Joseph Fisher, Christos Christodoulopoulos, and Andrea Pierleoni. 2022{\natexlab{b}}.
\newblock \href {https://doi.org/10.18653/v1/2022.naacl-industry.24} {{R}e{F}in{ED}: An efficient zero-shot-capable approach to end-to-end entity linking}.
\newblock In \emph{Proceedings of the 2022 Conference of the North American Chapter of the Association for Computational Linguistics: Human Language Technologies: Industry Track}, pages 209--220. Association for Computational Linguistics.

\bibitem[{Ayoola et~al.(2022{\natexlab{c}})Ayoola, Tyagi, Fisher, Christodoulopoulos, and Pierleoni}]{ayoola-etal-2022-refined}
Tom Ayoola, Shubhi Tyagi, Joseph Fisher, Christos Christodoulopoulos, and Andrea Pierleoni. 2022{\natexlab{c}}.
\newblock \href {https://doi.org/10.18653/v1/2022.naacl-industry.24} {{R}e{F}in{ED}: An efficient zero-shot-capable approach to end-to-end entity linking}.
\newblock In \emph{Proceedings of the 2022 Conference of the North American Chapter of the Association for Computational Linguistics: Human Language Technologies: Industry Track}, pages 209--220, Hybrid: Seattle, Washington + Online. Association for Computational Linguistics.

\bibitem[{Cai et~al.(2016)Cai, Zhang, and Wang}]{cai_bidirectional_2016}
Rui Cai, Xiaodong Zhang, and Houfeng Wang. 2016.
\newblock \href {https://doi.org/10.18653/v1/P16-1072} {Bidirectional {Recurrent} {Convolutional} {Neural} {Network} for {Relation} {Classification}}.
\newblock In \emph{Proceedings of the 54th {Annual} {Meeting} of the {Association} for {Computational} {Linguistics} ({Volume} 1: {Long} {Papers})}, pages 756--765, Berlin, Germany. Association for Computational Linguistics.

\bibitem[{Eberts and Ulges(2021)}]{eberts_end--end_2021}
Markus Eberts and Adrian Ulges. 2021.
\newblock \href {https://doi.org/10.18653/v1/2021.eacl-main.319} {An {End}-to-end {Model} for {Entity}-level {Relation} {Extraction} using {Multi}-instance {Learning}}.
\newblock In \emph{Proceedings of the 16th {Conference} of the {European} {Chapter} of the {Association} for {Computational} {Linguistics}: {Main} {Volume}}, pages 3650--3660, Online. Association for Computational Linguistics.

\bibitem[{Feng et~al.(2018)Feng, Huang, Zhao, Yang, and Zhu}]{feng_reinforcement_2018}
Jun Feng, Minlie Huang, Li~Zhao, Yang Yang, and Xiaoyan Zhu. 2018.
\newblock \href {https://doi.org/10.1609/aaai.v32i1.12063} {Reinforcement {Learning} for {Relation} {Classification} {From} {Noisy} {Data}}.
\newblock \emph{Proceedings of the AAAI Conference on Artificial Intelligence}, 32(1).

\bibitem[{Genest and Lapalme(2012)}]{genest2012fully}
Pierre-Etienne Genest and Guy Lapalme. 2012.
\newblock Fully abstractive approach to guided summarization.
\newblock In \emph{Proceedings of the 50th Annual Meeting of the Association for Computational Linguistics (Volume 2: Short Papers)}, pages 354--358.

\bibitem[{Han et~al.(2018)Han, Yu, Liu, Sun, and Li}]{han_hierarchical_2018}
Xu~Han, Pengfei Yu, Zhiyuan Liu, Maosong Sun, and Peng Li. 2018.
\newblock \href {https://doi.org/10.18653/v1/D18-1247} {Hierarchical {Relation} {Extraction} with {Coarse}-to-{Fine} {Grained} {Attention}}.
\newblock In \emph{Proceedings of the 2018 {Conference} on {Empirical} {Methods} in {Natural} {Language} {Processing}}, pages 2236--2245, Brussels, Belgium. Association for Computational Linguistics.

\bibitem[{Huguet~Cabot and Navigli(2021)}]{huguet-cabot-navigli-2021-rebel-relation}
Pere-Llu{\'\i}s Huguet~Cabot and Roberto Navigli. 2021.
\newblock \href {https://doi.org/10.18653/v1/2021.findings-emnlp.204} {{REBEL}: Relation extraction by end-to-end language generation}.
\newblock In \emph{Findings of the Association for Computational Linguistics: EMNLP 2021}, pages 2370--2381, Punta Cana, Dominican Republic. Association for Computational Linguistics.

\bibitem[{Josifoski et~al.(2022)Josifoski, De~Cao, Peyrard, Petroni, and West}]{josifoski-etal-2022-genie}
Martin Josifoski, Nicola De~Cao, Maxime Peyrard, Fabio Petroni, and Robert West. 2022.
\newblock \href {https://doi.org/10.18653/v1/2022.naacl-main.342} {{G}en{IE}: Generative information extraction}.
\newblock In \emph{Proceedings of the 2022 Conference of the North American Chapter of the Association for Computational Linguistics: Human Language Technologies}, pages 4626--4643, Seattle, United States. Association for Computational Linguistics.

\bibitem[{Kingma and Ba(2015)}]{adam}
Diederik~P. Kingma and Jimmy Ba. 2015.
\newblock \href {http://arxiv.org/abs/1412.6980} {Adam: {A} method for stochastic optimization}.
\newblock In \emph{3rd International Conference on Learning Representations, {ICLR} 2015, San Diego, CA, USA, May 7-9, 2015, Conference Track Proceedings}.

\bibitem[{Lee et~al.(2018)Lee, He, and Zettlemoyer}]{lee_higher-order_2018}
Kenton Lee, Luheng He, and Luke Zettlemoyer. 2018.
\newblock \href {https://doi.org/10.18653/v1/N18-2108} {Higher-{Order} {Coreference} {Resolution} with {Coarse}-to-{Fine} {Inference}}.
\newblock In \emph{Proceedings of the 2018 {Conference} of the {North} {American} {Chapter} of the {Association} for {Computational} {Linguistics}: {Human} {Language} {Technologies}, {Volume} 2 ({Short} {Papers})}, pages 687--692, New Orleans, Louisiana. Association for Computational Linguistics.

\bibitem[{Liu et~al.(2019)Liu, Ott, Goyal, Du, Joshi, Chen, Levy, Lewis, Zettlemoyer, and Stoyanov}]{liu_roberta_2019}
Yinhan Liu, Myle Ott, Naman Goyal, Jingfei Du, Mandar Joshi, Danqi Chen, Omer Levy, Mike Lewis, Luke Zettlemoyer, and Veselin Stoyanov. 2019.
\newblock \href {https://doi.org/10.48550/arXiv.1907.11692} {{RoBERTa}: {A} {Robustly} {Optimized} {BERT} {Pretraining} {Approach}}.

\bibitem[{Liu et~al.(2018)Liu, Zhang, Liang, Ji, and McGuinness}]{liu_seq2rdf_2018}
Yue Liu, Tongtao Zhang, Zhicheng Liang, Heng Ji, and Deborah~L. McGuinness. 2018.
\newblock \href {http://www.scopus.com/inward/record.url?scp=85055312888&partnerID=8YFLogxK} {{Seq2rDF}: 2018 {ISWC} {Posters} and {Demonstrations}, {Industry} and {Blue} {Sky} {Ideas} {Tracks}, {ISWC}-{P} and {D}-{Industry}-{BlueSky} 2018}.
\newblock \emph{CEUR Workshop Proceedings}, 2180.

\bibitem[{Mesquita et~al.(2019)Mesquita, Cannaviccio, Schmidek, Mirza, and Barbosa}]{mesquita_knowledgenet_2019}
Filipe Mesquita, Matteo Cannaviccio, Jordan Schmidek, Paramita Mirza, and Denilson Barbosa. 2019.
\newblock \href {https://doi.org/10.18653/v1/D19-1069} {{KnowledgeNet}: {A} {Benchmark} {Dataset} for {Knowledge} {Base} {Population}}.
\newblock In \emph{Proceedings of the 2019 {Conference} on {Empirical} {Methods} in {Natural} {Language} {Processing} and the 9th {International} {Joint} {Conference} on {Natural} {Language} {Processing} ({EMNLP}-{IJCNLP})}, pages 749--758, Hong Kong, China. Association for Computational Linguistics.

\bibitem[{Miwa and Bansal(2016)}]{miwa_end--end_2016}
Makoto Miwa and Mohit Bansal. 2016.
\newblock \href {https://doi.org/10.18653/v1/P16-1105} {End-to-{End} {Relation} {Extraction} using {LSTMs} on {Sequences} and {Tree} {Structures}}.
\newblock In \emph{Proceedings of the 54th {Annual} {Meeting} of the {Association} for {Computational} {Linguistics} ({Volume} 1: {Long} {Papers})}, pages 1105--1116, Berlin, Germany. Association for Computational Linguistics.

\bibitem[{Miwa and Sasaki(2014)}]{miwa_modeling_2014}
Makoto Miwa and Yutaka Sasaki. 2014.
\newblock \href {https://doi.org/10.3115/v1/D14-1200} {Modeling {Joint} {Entity} and {Relation} {Extraction} with {Table} {Representation}}.
\newblock In \emph{Proceedings of the 2014 {Conference} on {Empirical} {Methods} in {Natural} {Language} {Processing} ({EMNLP})}, pages 1858--1869, Doha, Qatar. Association for Computational Linguistics.

\bibitem[{Muhammad et~al.(2020)Muhammad, Kearney, Gamble, Coenen, and Williamson}]{muhammad_open_2020}
Iqra Muhammad, Anna Kearney, Carrol Gamble, Frans Coenen, and Paula Williamson. 2020.
\newblock \href {https://doi.org/10.1007/978-3-030-59028-4_10} {Open {Information} {Extraction} for {Knowledge} {Graph} {Construction}}.
\newblock In \emph{Database and {Expert} {Systems} {Applications}}, Communications in {Computer} and {Information} {Science}, pages 103--113, Cham. Springer International Publishing.

\bibitem[{Nasar et~al.(2021)Nasar, Jaffry, and Malik}]{nasar_named_2021}
Zara Nasar, Syed~Waqar Jaffry, and Muhammad~Kamran Malik. 2021.
\newblock \href {https://doi.org/10.1145/3445965} {Named {Entity} {Recognition} and {Relation} {Extraction}: {State}-of-the-{Art}}.
\newblock \emph{ACM Computing Surveys}, 54(1):20:1--20:39.

\bibitem[{Pawar et~al.(2017)Pawar, Bhattacharyya, and Palshikar}]{pawar_end--end_2017}
Sachin Pawar, Pushpak Bhattacharyya, and Girish Palshikar. 2017.
\newblock \href {https://aclanthology.org/E17-1077} {End-to-end {Relation} {Extraction} using {Neural} {Networks} and {Markov} {Logic} {Networks}}.
\newblock In \emph{Proceedings of the 15th {Conference} of the {European} {Chapter} of the {Association} for {Computational} {Linguistics}: {Volume} 1, {Long} {Papers}}, pages 818--827, Valencia, Spain. Association for Computational Linguistics.

\bibitem[{Provatorova et~al.(2021)Provatorova, Bhargav, Vakulenko, and Kanoulas}]{shadowlink}
Vera Provatorova, Samarth Bhargav, Svitlana Vakulenko, and Evangelos Kanoulas. 2021.
\newblock \href {https://doi.org/10.18653/v1/2021.emnlp-main.820} {Robustness evaluation of entity disambiguation using prior probes: the case of entity overshadowing}.
\newblock In \emph{Proceedings of the 2021 Conference on Empirical Methods in Natural Language Processing}, pages 10501--10510, Online and Punta Cana, Dominican Republic. Association for Computational Linguistics.

\bibitem[{Raiman and Raiman(2018)}]{Raiman2018DeepTypeME}
Jonathan Raiman and O.~Raiman. 2018.
\newblock Deeptype: Multilingual entity linking by neural type system evolution.
\newblock In \emph{AAAI}.

\bibitem[{Ramshaw and Marcus(1995)}]{ramshaw-marcus-1995-text}
Lance Ramshaw and Mitch Marcus. 1995.
\newblock \href {https://aclanthology.org/W95-0107} {Text chunking using transformation-based learning}.
\newblock In \emph{Third Workshop on Very Large Corpora}.

\bibitem[{Sui et~al.(2021)Sui, Wang, Chen, Liu, Zhao, and Bi}]{sui_set_2021}
Dianbo Sui, Chenhao Wang, Yubo Chen, Kang Liu, Jun Zhao, and Wei Bi. 2021.
\newblock \href {https://doi.org/10.18653/v1/2021.emnlp-main.760} {Set {Generation} {Networks} for {End}-to-{End} {Knowledge} {Base} {Population}}.
\newblock In \emph{Proceedings of the 2021 {Conference} on {Empirical} {Methods} in {Natural} {Language} {Processing}}, pages 9650--9660, Online and Punta Cana, Dominican Republic. Association for Computational Linguistics.

\bibitem[{Taill{\'e} et~al.(2021)Taill{\'e}, Guigue, Scoutheeten, and Gallinari}]{taille-etal-2021-separating}
Bruno Taill{\'e}, Vincent Guigue, Geoffrey Scoutheeten, and Patrick Gallinari. 2021.
\newblock \href {https://doi.org/10.18653/v1/2021.emnlp-main.816} {Separating retention from extraction in the evaluation of end-to-end {R}elation {E}xtraction}.
\newblock In \emph{Proceedings of the 2021 Conference on Empirical Methods in Natural Language Processing}, pages 10438--10449, Online and Punta Cana, Dominican Republic. Association for Computational Linguistics.

\bibitem[{Tan et~al.(2022)Tan, Xu, Bing, Ng, and Aljunied}]{tan2022revisiting}
Qingyu Tan, Lu~Xu, Lidong Bing, Hwee~Tou Ng, and Sharifah~Mahani Aljunied. 2022.
\newblock \href {https://arxiv.org/abs/2205.12696} {Revisiting docred – addressing the false negative problem in relation extraction}.
\newblock In \emph{Proceedings of EMNLP}.

\bibitem[{Trisedya et~al.(2019)Trisedya, Weikum, Qi, and Zhang}]{trisedya_neural_2019}
Bayu~Distiawan Trisedya, Gerhard Weikum, Jianzhong Qi, and Rui Zhang. 2019.
\newblock \href {https://doi.org/10.18653/v1/P19-1023} {Neural {Relation} {Extraction} for {Knowledge} {Base} {Enrichment}}.
\newblock In \emph{Proceedings of the 57th {Annual} {Meeting} of the {Association} for {Computational} {Linguistics}}, pages 229--240, Florence, Italy. Association for Computational Linguistics.

\bibitem[{Verlinden et~al.(2021)Verlinden, Zaporojets, Deleu, Demeester, and Develder}]{verlinden_injecting_2021}
Severine Verlinden, Klim Zaporojets, Johannes Deleu, Thomas Demeester, and Chris Develder. 2021.
\newblock \href {https://doi.org/10.18653/v1/2021.findings-acl.171} {Injecting {Knowledge} {Base} {Information} into {End}-to-{End} {Joint} {Entity} and {Relation} {Extraction} and {Coreference} {Resolution}}.
\newblock In \emph{Findings of the {Association} for {Computational} {Linguistics}: {ACL}-{IJCNLP} 2021}, pages 1952--1957, Online. Association for Computational Linguistics.

\bibitem[{Wang et~al.(2020)Wang, Hu, Cao, and Sun}]{wang_global--local_2020}
D.~Wang, Wei Hu, E.~Cao, and Weijian Sun. 2020.
\newblock \href {https://doi.org/10.18653/v1/2020.emnlp-main.303} {Global-to-{Local} {Neural} {Networks} for {Document}-{Level} {Relation} {Extraction}}.
\newblock \emph{EMNLP}.

\bibitem[{Wolf et~al.(2019)Wolf, Debut, Sanh, Chaumond, Delangue, Moi, Cistac, Rault, Louf, Funtowicz, and Brew}]{huggingface}
Thomas Wolf, Lysandre Debut, Victor Sanh, Julien Chaumond, Clement Delangue, Anthony Moi, Pierric Cistac, Tim Rault, R'emi Louf, Morgan Funtowicz, and Jamie Brew. 2019.
\newblock Huggingface's transformers: State-of-the-art natural language processing.
\newblock \emph{ArXiv}, abs/1910.03771.

\bibitem[{Xu et~al.(2021)Xu, Wang, Lyu, Zhu, and Mao}]{xu_entity_2021}
Benfeng Xu, Quan Wang, Yajuan Lyu, Yong Zhu, and Zhendong Mao. 2021.
\newblock \href {https://doi.org/10.1609/aaai.v35i16.17665} {Entity {Structure} {Within} and {Throughout}: {Modeling} {Mention} {Dependencies} for {Document}-{Level} {Relation} {Extraction}}.
\newblock \emph{Proceedings of the AAAI Conference on Artificial Intelligence}, 35(16):14149--14157.
\newblock Number: 16.

\bibitem[{Xu and Choi(2022)}]{xu_modeling_2022}
Liyan Xu and Jinho~D. Choi. 2022.
\newblock \href {https://doi.org/10.48550/arXiv.2205.01909} {Modeling {Task} {Interactions} in {Document}-{Level} {Joint} {Entity} and {Relation} {Extraction}}.
\newblock ArXiv:2205.01909 [cs].

\bibitem[{Yao and Van~Durme(2014)}]{yao_information_2014}
Xuchen Yao and Benjamin Van~Durme. 2014.
\newblock \href {https://doi.org/10.3115/v1/P14-1090} {Information {Extraction} over {Structured} {Data}: {Question} {Answering} with {Freebase}}.
\newblock In \emph{Proceedings of the 52nd {Annual} {Meeting} of the {Association} for {Computational} {Linguistics} ({Volume} 1: {Long} {Papers})}, pages 956--966, Baltimore, Maryland. Association for Computational Linguistics.

\bibitem[{Yao et~al.(2019)Yao, Ye, Li, Han, Lin, Liu, Liu, Huang, Zhou, and Sun}]{yao_docred_2019}
Yuan Yao, Deming Ye, Peng Li, Xu~Han, Yankai Lin, Zhenghao Liu, Zhiyuan Liu, Lixin Huang, Jie Zhou, and Maosong Sun. 2019.
\newblock \href {http://arxiv.org/abs/1906.06127} {{DocRED}: {A} {Large}-{Scale} {Document}-{Level} {Relation} {Extraction} {Dataset}}.
\newblock \emph{arXiv:1906.06127 [cs]}.
\newblock ArXiv: 1906.06127.

\bibitem[{Zaporojets et~al.(2021)Zaporojets, Deleu, Develder, and Demeester}]{zaporojets_dwie_2021}
Klim Zaporojets, Johannes Deleu, Chris Develder, and Thomas Demeester. 2021.
\newblock \href {https://doi.org/10.1016/j.ipm.2021.102563} {{DWIE}: {An} entity-centric dataset for multi-task document-level information extraction}.
\newblock \emph{Information Processing \& Management}, 58(4):102563.

\bibitem[{Zeng et~al.(2020)Zeng, Xu, Chang, and Li}]{zeng_double_2020}
Shuang Zeng, Runxin Xu, Baobao Chang, and Lei Li. 2020.
\newblock \href {https://doi.org/10.18653/v1/2020.emnlp-main.127} {Double {Graph} {Based} {Reasoning} for {Document}-level {Relation} {Extraction}}.
\newblock In \emph{Proceedings of the 2020 {Conference} on {Empirical} {Methods} in {Natural} {Language} {Processing} ({EMNLP})}, pages 1630--1640, Online. Association for Computational Linguistics.

\bibitem[{Zhang et~al.(2021)Zhang, Chen, Xie, Deng, Tan, Chen, Huang, Si, and Chen}]{zhang_document-level_2021}
Ningyu Zhang, Xiang Chen, Xin Xie, Shumin Deng, Chuanqi Tan, Mosha Chen, Fei Huang, Luo Si, and Huajun Chen. 2021.
\newblock \href {http://arxiv.org/abs/2106.03618} {Document-level {Relation} {Extraction} as {Semantic} {Segmentation}}.
\newblock \emph{arXiv:2106.03618 [cs]}.
\newblock ArXiv: 2106.03618.

\end{thebibliography}

\clearpage

\appendix

\label{sec:appendix}

\section{Coref clustering}
\label{sec:coref_app}
\begin{figure*}[htp]
\centering
\includegraphics[width=1\linewidth]{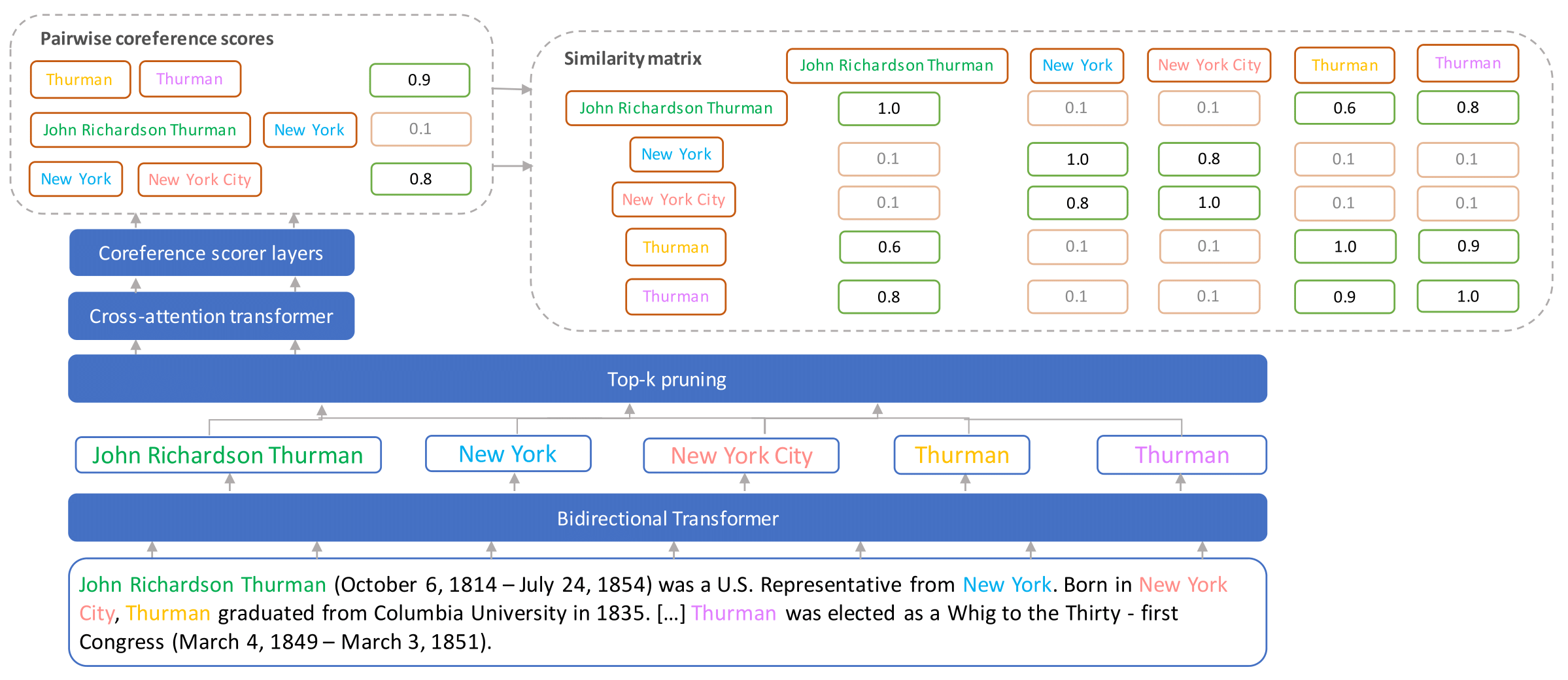}
\caption{Architecture of the Coreference Resolution module}
\label{fig:COREF_model_architecture}
\end{figure*}
We detail the different approaches used for coreference clustering in the following sections.

\subsection{Entity Linking} 
We use entity disambiguation for predicting an identifier for each mention, and then cluster mentions which have the same identifier. This approach relies on external knowledge. Also, this approach necessitates performing entity disambiguation to obtain the identifiers, which may not always be part of the task of interest, e.g., RE does not require ED.

\subsection{Greedy approach}
Let's consider a set of mentions to cluster $\left( m_i\right)_{1\le i \le N}$. The greedy approach comprises two stages: first, forming a similarity matrix $S \in \mathbb{R}^{N \times N}$ from the pairwise scores, and second, forming the cluster $(C_i)_{i}$. The model is trained on the pairwise scores only. The clusters are then defined as follows:

\begin{equation}
\begin{split}
    C_i \vcentcolon= \left\{ m_j : \forall j \in [|1, N|] \right. \\
    \text{ such that } S_{i, j} >  t  \text{ and }\\
    \left. m_j \notin C_k \text{ for } 1 \le k \le i-1 \right\}
\end{split}
\end{equation}

where $t \in [0, 1]$ is the coreference threshold and $S_{i, i} = 1\ \forall i \in [|1, N|]$. This approach iteratively considers each mention $m_i$ and constructs a cluster based on the coreference scores between $m_i$ and all other valid mentions, where a valid mention is one that has not yet been assigned to a cluster. Notably, each mention span is allocated to only one cluster. However, it's crucial to acknowledge that the hard-metric constraint implies that any absent mention within a cluster renders the entire cluster invalid. \\

Hence, we explore an alternative approach that relaxes the constraint of a mention belonging to only one cluster. This variant, termed the \emph{Greedy approach (multiple-clusters)}, allows mentions to be assigned to multiple clusters simultaneously. Each cluster is then defined as follows:

\begin{equation}
    C_i \vcentcolon= \left\{ m_j : \forall j \in [|1, N|] \text{ such that } S_{i, j} > t\right\}
\end{equation}

\subsection{Agglomerative Clustering}
The agglomerative clustering approach also relies on forming a similarity matrix, see Figure~\ref{fig:COREF_model_architecture}. The model is trained to predict pairwise coreference scores rather than directly predicting the clusters. Put simply, the coreference resolution component of our model is optimized for predicting a similarity matrix. Then, the second stage exploits that matrix to form the clusters. The distance threshold was chosen experimentally and we did not perform hyperparameter tuning to optimize it. The coreference performance may be further improved by including the threshold in the training.

\begin{table}[t]
    \centering
    \small

    \begin{tabular}{l c ccc}
    \hline
    \textbf{COREF methods} & & \textbf{P} & \textbf{R} & \textbf{F1} \\
    \hline
    Greedy                         & {} & 0.89 & 0.9 & 0.9 \\
    Greedy (multiple-clusters)     & {} & 0.88 & 0.9 & 0.89 \\
    EL-based                       & {} &  0.88 & 0.89  &  0.89 \\
    Complete linkage               & {} & 0.89 & 0.9 & 0.9 \\
    Average linkage                & {} & 0.9 & 0.9 & 0.9\\
    \hline
    \end{tabular}
    \caption{Coref evaluation using different approaches}
    \label{tab:Coref_approaches_DocRED_dev}
\end{table}

\section{Training Details}
\label{sec:training-details}

REXEL uses Hugging Face implementation of RoBERTa \cite{huggingface} and
the model is optimised using Adam \cite{adam} with a linear learning rate schedule. Our main hyperparameters are represented in Table \ref{tab:hyperparams}. Due to the high computational cost of training the model, we did not conduct an extensive hyperparameter search. Training across datasets took approximately 24 hours on average on a single machine with 1 V100 GPU. REXEL has approximately 284M parameters in its architecture setup.

\begin{table}[h]
\resizebox{0.9\linewidth}{!}{%
\begin{tabular}{@{}ll@{}}
\toprule
\textbf{Hyperparameter} & \textbf{Value} \\ \midrule
learning rate           & 5e-5           \\
batch size              & 2             \\
max sequence length     & 510            \\
dropout                 & 0.1           \\
RC threshold & 0.2 \\
description embeddings dim.    & 300            \\ 
\# training epochs                & 150             \\
\# candidates           & 30             \\
\# wikidata entity types      & 1400            \\ 
mention transformer init.  & roberta-base \\
\# mention encoder layers     & 12            \\
description transformer init.  & roberta-base \\
\# description encoder layers     & 2            \\ 
\# RC encoder layers & 4 \\
RC coarse-to-fine k & 2000 \\ 
\# description tokens     & 32            \\
$\lambda_1$, $\lambda_2$, $\lambda_3$, $\lambda_4$, $\lambda_5$  & (0.1, 0.005, 0.1, 0.02, 0.775)\\

\bottomrule
\end{tabular}
}
\caption{Our model hyperparameters}
\label{tab:hyperparams}
\end{table}

\section{Datasets}
\label{sec:datasets_appendix}

\subsection{DocRED and DWIE}

The DocRED dataset was constructed from Wikipedia documents, whereas DWIE was constructed from news articles.  DocRED and DWIE both comprise document-level and sentence-level facts, and they are both annotated at entity-level, i.e., facts are reported between entity clusters made of several mentions, which motivates the additional coreference resolution step for extracting relations. Also, they both require different types of reasoning to extract triples living across multiple sentences (e.g, pattern recognition, logical or common-sense reasoning). We report some statistics on these dataset in Table~\ref{tab:dataset_stats}. Another similarity is that both datasets have a class-imbalance problem, which increases the complexity of the RC task.  More precisely, 10 relations account for about 60\% of the facts in DocRED, while the 10 most frequent relations account for more than 75\% of the facts for DWIE. In addition, DocRED-E2E contains some duplicate annotations, which we remove at evaluation stage following the convention introduced by \cite{eberts_end--end_2021}. Likewise, DWIE contains some spurious empty clusters (see Table~\ref{tab:stats_mentions_per_cluster}), which we remove with their associated triples following the setting adopted by \cite{xu_modeling_2022}.

\begin{table}
\centering
\small
\begin{tabular}{lccc}
\toprule
& DocRED & DocRED-E2E & DWIE \\
\midrule
\# Documents      & 5051 & 4008 & 802\\
\# Entities/doc   & 19.5 & 19.4 & 28.3 \\
\# Facts/doc      &  13.2 & 12.5 & 27 \\
\# Entity types   & 6 & 6 & 311 \\
\# Relations      & 96 & 96  & 65 \\
\bottomrule
\end{tabular}
\caption{Some statistics for DocRED, DocRED-E2E and DWIE. \emph{\# Entities/doc} and \emph{\# Facts/doc} refer respectively to the averaged number of entities and facts per document.}
    \label{tab:dataset_stats}
\end{table}

\begin{table}
    \centering
    \small
    \begin{tabular}{lcc}
    \toprule
     \# Mentions/Entity  & DocRED-E2E (\%) & DWIE (\%) \\
    \midrule
    0 & 0 & 5.3 \\
    1 & 81.7 & 62.9 \\
    2 & 11.1 & 14.4\\
    3 & 3.6  & 6.1 \\
    > 4 & 3.6  & 11.3 \\
    \bottomrule
    \end{tabular}
    \caption{Proportion of mentions per entity cluster in DocRED-E2E and DWIE.}
    \label{tab:stats_mentions_per_cluster}
\end{table}

\subsection{DocRED-IE}

To facilitate future works on DocIE, we release \textit{DocRED-IE}, an extension of the DocRED \cite{yao_docred_2019} dataset further equipped with entity links, making it the second dataset to support DocIE evaluation, thereby facilitating future research on document-level closed information extraction. DocRED-IE allows for training and evaluation in a multitask setting encompassing mention detection, entity typing, coreference resolution, document-level relation classification, and entity linking, along with any combination thereof in a joint setting, such as the end-to-end RE task and DocIE. \\

DocRED-IE builds on the end-to-end DocRED release introduced in \cite{eberts_end--end_2021} (DocRED-E2E). We employ a state-of-the-art entity linking model \cite{ayoola-etal-2022-refined} to populate each mention in DocRED-E2E. Statistics of the DocRED-IE dataset are shown in Table~\ref{tab:docredie_stats}.

\begin{table}
\centering
\small
\begin{tabular}{lccc}
\toprule
& Train & Dev & Test\\
\midrule
\# Documents        & 3008 & 300 & 700 \\
\# Entities         & 58708 & 5805 & 13594\\
\# Entities linked  & 45874 & 4025 & 10191 \\
\# Facts            & 37486 & 3678 & 8787\\
\# Entity types     & 6  & 6  & 6 \\
\# Relations        & 96 & 96 & 96 \\
\bottomrule
\end{tabular}
\caption{Some statistics for DocRED-IE.}
    \label{tab:docredie_stats}
\end{table}

\subsection{Dataset Licenses}
The DWIE dataset \cite{zaporojets_dwie_2021} and the code has been released under GNU GPLv3 license \footnote{https://github.com/klimzaporojets/DWIE/blob/master/LICENSE}. Both the DocRED-E2E \footnote{https://github.com/lavis-nlp/jerex/blob/main/LICENSE} dataset \cite{eberts_end--end_2021} and DocRED-IE are released under MIT licence.

\end{document}